\documentclass[10pt,twocolumn,letterpaper]{article}

\usepackage{wacv}
\usepackage{times}
\usepackage{epsfig}
\usepackage{graphicx}
\usepackage{amsmath}
\usepackage{amssymb}
\usepackage{colortbl}
\usepackage{multirow}
\usepackage{booktabs}

\def\wacvPaperID{1188} % Enter the WACV Paper ID here

\wacvfinalcopy % *** Uncomment this line for the final submission

\ifwacvfinal
\def\assignedStartPage{9876} % *** Enter the assigned starting page number (instead of 9876)
\fi

\ifwacvfinal
\else
\fi

\begin{document}

%%%%%%%%% TITLE
\title{Mutual Information Maximization on Disentangled Representations for Differential Morph Detection}

\author{Sobhan Soleymani, Ali Dabouei, Fariborz Taherkhani, Jeremy Dawson, Nasser M. Nasrabadi\\
West Virginia University\\
{\tt\small \{ssoleyma, ad0046, ft0009\}@mix.wvu.edu, \{jeremy.dawson,
nasser.nasrabadi\}@mail.wvu.edu}
% \and
}

\maketitle
%\thispagestyle{empty}

%%%%%%%%% ABSTRACT
\begin{abstract}
In this paper, we present a novel differential morph detection framework, utilizing landmark and appearance disentanglement. In our framework, the face image is represented in the embedding domain using two disentangled but complementary representations. The network is trained by triplets of face images, in which the intermediate image inherits the landmarks from one image and the appearance from the other image. This initially trained network is further trained for each dataset using contrastive representations. We demonstrate that, by employing appearance and landmark disentanglement, the proposed framework can provide state-of-the-art differential morph detection performance. This functionality is achieved by the using distances in landmark, appearance, and ID domains. The performance of the proposed framework is evaluated using three morph datasets generated with different methodologies. 
\end{abstract}

%%%%%%%%% BODY TEXT
\section{Introduction}

The main goal of biometric systems is automated recognition of individuals based on their unique biological and behavioral characteristics~\cite{scherhag2017biometric}. The human face is widely accepted as a means of biometric authentication. Although, the uniqueness of face images and user convenience of face recognition systems have resulted in their popularity, morphed face images have shown to pose a severe threat to them. This is because the main objective of morph attacks is to purposefully alter or obfuscate the unique correspondence between probe and gallery images~\cite{scherhag2019face}. The result of a morph attack is a face image which matches the probe images corresponding to two different face images. Therefore, the detection of morph images plays a major role in providing reliable face recognition.

\begin{figure}
    \centering
    \includegraphics[width=230pt]{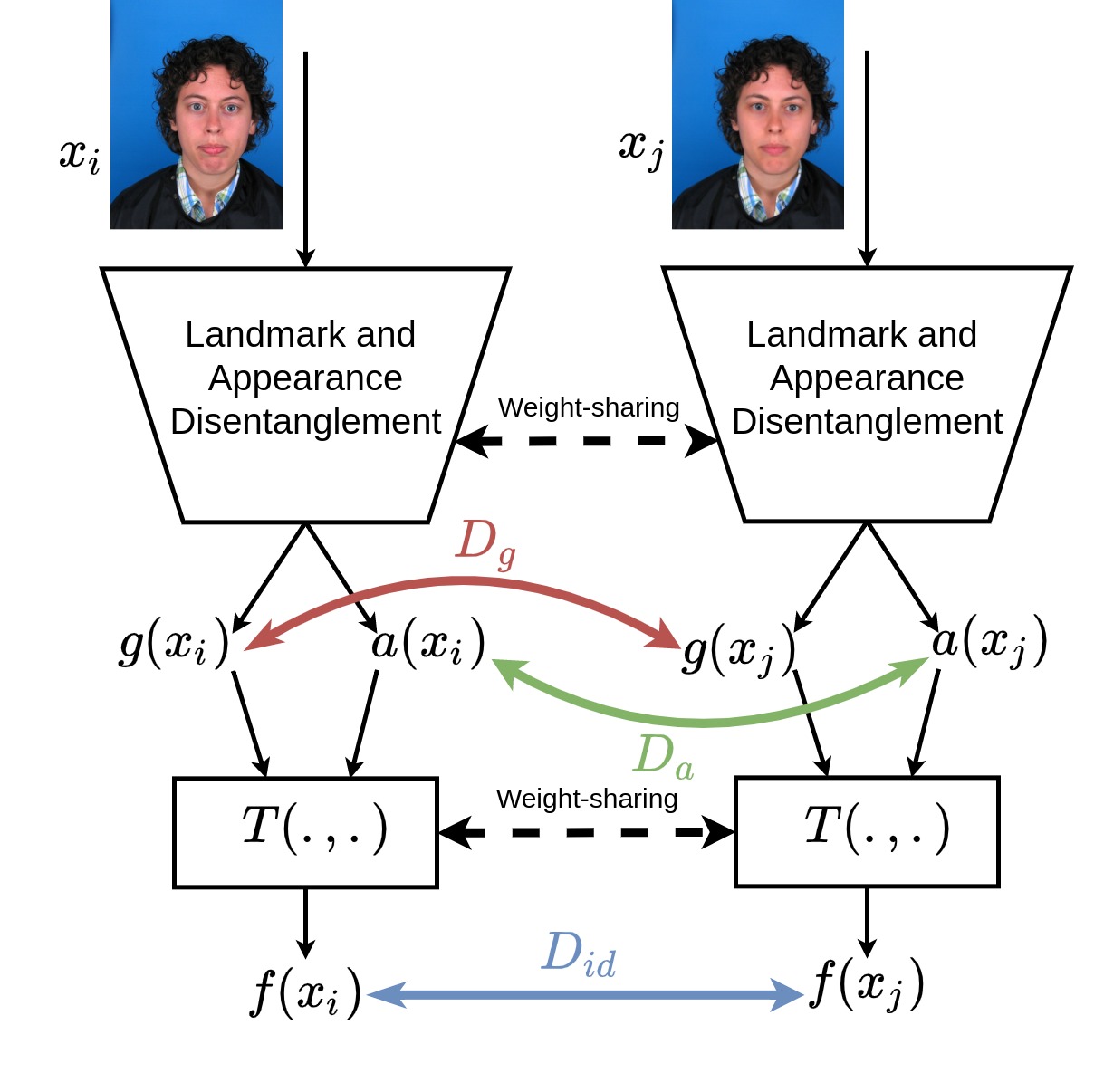}
    \caption{Trusted probe image, $x_i$, and image in question, $x_j$, are disentangled into landmark and appearance representations, using the disentanglement network trained on triplet of face images. In these triplets, the constructed intermediate face image inherits landmarks and the appearance from two different face images. Landmark, appearance, and ID representations are utilized to make the decision about the image in question.}
    \label{fig:firstfigure}
\end{figure}

The majority of morph generation frameworks focus on altering the position of the facial landmarks. These frameworks mainly utilize three steps: correspondence, warping, and blending. The first step aims to detect the corresponding landmarks of both the images. These sets of landmarks are then utilized to warp the images toward each other, {\it e.g.,} considering the landmarks of the morph image as the pairwise average of two face images. Finally, textures from the two images are combined either over the entire face image~\cite{scherhag2020deep} or face patches~\cite{makrushin2017automatic}. Another trend of morph generation considers Generative Adversarial Networks (GANs) to construct images that can be matched with the two source images, such as AliGAN~\cite{dumoulin2016adversarially,damer2018morgan} and StyleGAN~\cite{karras2019style,venkatesh2020can}. 

The face morphing algorithms can affect the face image in two broad aspects. First, they alter the position of the landmarks. On the other hand, they modify the appearance of the face image by either blending two source images or generating samples using generative models. Although appearance corresponds to the soft biometrics of a subject which are not necessarily unique, such as ethnicity, hair color, and gender, it can still be interpreted to distinguish between face images with similar soft biometrics such as differences in the texture of the face images. However, deep differential morph detection frameworks focus on distinguishing the samples based on the ID information. Our proposed differential morph detection framework investigates both the locations of the landmarks and the appearance of the face image. Therefore, this approach restricts the attacker's morphing capability by studying both the changes resulted from altering the landmarks as well as modification in soft biometrics and texture information. 

As presented in Figure~\ref{fig:firstfigure}, our proposed framework learns the disentangled representations for the landmarks and the appearance of a face image. While these representations are practically shown to be sufficient for face recognition~\cite{dabouei2020boosting}, the proposed training setup ensures that the mutual information between representations of the real images from a subject is maximized. In this paper, we make the following contributions: i) we construct triplets of images in which an intermediate image inherits the landmarks from one image and the appearance from the other image, ii) these triplets are considered to train a disentangling network which provides disentangled representations for landmarks and face appearance, and iii) we train specific networks for each morph dataset by learning contrastive representations through maximizing the mutual information between real images from each subject.

\section{Related Works}
\subsection{Facial Morphing}
Facial morphing studies the possibility of creating artificial biometric samples which resemble the biometric information of two or more individuals~\cite{scherhag2019face}. Morph images can be generated with little technical experience using tools available on the internet and mobile platforms~\cite{scherhag2019face}. The overall purpose of face morphing is to generate a face image that will be verified against samples of two or more subjects in automated face recognition
systems. One of the first efforts to study the generation of a morph image from two source images~\cite{ferrara2014magic} has concluded that geometric alterations and digital beautification can cause an increase in the possibility of fooling  recognition systems. Morph generation techniques can roughly be categorized into landmark-based~\cite{makrushin2017automatic,seibold2017detection,seibold2018accurate} and generative models~\cite{damer2018morgan,sharif2017adversarial}. Landmark-based frameworks focus on detecting the landmarks in both the images, translating these points toward each other, and blending the two face images.  On the other hand, inspired by a learned inference model~\cite{dumoulin2016adversarially}, Morgan~\cite{damer2018morgan} presents a face morphing attack based on automatic image generation using a GAN framework.

\subsection{Morph Detection}
Morph detection can be categorized into two main approaches~\cite{scherhag2020deep}: single image morph detection and differential morph detection. Single image morph detection studies the possibility of detecting the morph image in the absence of a reference image. On the other hand, differential morph detection leverages the information extracted from a real image corresponding to the subject. Texture descriptors are the main feature extraction models for single image morph detection~\cite{wandzik2018morphing,spreeuwers2018towards,scherhag2018morph,scherhag2017vulnerability,ramachandra2019towards}. Recently, deep learning models have also been considered for this purpose~\cite{seibold2018accurate,seibold2017detection,raja2017transferable}. The models mentioned can also be employed for differential morph detection when the extracted feature from the two images are compared~\cite{scherhag2018detecting,scherhag2018towards,damer2018detecting}. Another trend of work for differential morph detection considers that subtracting the trusted image from the image in question should increase the classification score of the resulting image for one of the probe subjects~\cite{ferrara2017face,ferrara2018face,peng2019fd}.

\subsection{Representation Disentanglement}
The geometry of landmarks and visual appearance are the two main characteristics of the face that can be utilized for face recognition. Initially, the geometry of hand-crafted face landmarks were basis for face recognition~\cite{galton1889personal}. Neural network approaches have provided state-of-the-art face recognition performance, with several deep models using the location of landmarks for varying face recognition purposes~\cite{iranmanesh2020robust,dabouei2019fast}. On the other hand, the effect of appearance in face recognition is widely studied, including soft biometrics such as gender, age, ethnicity, and hair color~\cite{gonzalez2018facial,galiyawala2020person}. Recently, an unsupervised approach using a coupled autoencoder model for disentangling the appearance and geometry of face images was developed~\cite{shu2018deforming}. In this framework, each autoencoder learns the geometry or appearance representation of the face, while the reconstruction loss is considered as the supervision for disentangling. Another similar work~\cite{xing2018deformable} has incorporated variational autoencoders to improve the disentangling. Another recent generative model~\cite{kazemi2019style} presents an unsupervised algorithm for training GANs that learns the disentangled style and content representations of the data.

\subsection{Mutual Information and Deep Learning}
Among the first works that studied the application of mutual information in deep learning, \cite{nowozin2016f} showed that GAN training loss can be recovered by minimizing the estimated divergence between the generated and true data distributions. The authors in~\cite{belghazi2018mutual} expanded the mutual information maximization techniques to estimate the mutual information between two random variables via a neural network.
The authors in~\cite{cakir2017mihash} and~\cite{cakir2019hashing} used mutual information to quantify the separation of distributions of positive and negative pairings in learning binary hash codes. The authors in~\cite{kemertas2020rankmi} introduced RankMI algorithm,an information-theoretic loss function and a training algorithm for deep representation learning for image retrieval. The authors in contrastive representation distillation~\cite{tian2019contrastive} proposed a contrastive-based objective function for transferring knowledge between deep networks. The authors in~\cite{bachman2019learning} propose an approach to self-supervised representation learning based on maximizing mutual information between features extracted from multiple views of a shared context.

\section{Proposed Framework}
Our proposed differential morph detection framework resonates with the morph generation frameworks in which the the landmarks of the real image are translated to landmarks of the target face image~\cite{makrushin2017automatic} or image generation by generative adversarial networks~\cite{damer2018morgan}. Disentangling appearance and landmark information has shown to be a powerful tool for face recognition~\cite{dabouei2020boosting}. These two domains provide the majority of the information content for differential morph detection as well. We aim to study the possibility of detecting the morph image based on its differences with the trusted image in both landmark and appearance domains. Therefore, to train our framework, we construct samples that inherit the appearance and landmarks from different samples. Then, we train a network that can disentangle these two types of information~\cite{dabouei2020boosting}. This framework is then trained for differential morph detection by maximizing the mutual information between representations of genuine pairs. 

\subsection{Landmark and Appearance Triplets}

The first step in our proposed training consists of generating face images that inherit appearance from one image and landmarks from the other image. Then, these triplets of face images are used to train two deep networks. The first network aims to represent the appearance of the face image and the second network extracts the landmark information.   
The supervision for disentangling appearance and landmarks of faces is provided by constructing triplets of face images. Each triplet consists of two real face images from two different IDs. For convenience we denote these images as appearance image, $x_i$, landmark image, $x'_i$, and an intermediate face image generated using the appearance of the first face image and the landmarks of the second face image, $\widehat{x_i}$. To construct this intermediate face image, we translate the landmarks of the appearance image to the landmark image. 

For this purpose, let $x_i$ be a face image noted as an appearance image belonging to the class $y_i$ and the set $l_i$ describe the locations of its $K$ landmarks. We find another face image $x'_i$ from a different class corresponding to the closest set of landmarks $l'_i$ as the landmark set. The distance between the sets of landmarks is calculated in terms of $L_\infty$, to assure that $x_i$ and $x'_i$ have similar structures in order to minimize the distortion caused by the spatial transformation in the next step. 

We use the thin plate spline (TPS) algorithm~\cite{bookstein1989principal} to transfer the landmarks of the appearance face image to the landmarks of $x'_i$ as: 

\begin{equation}
\widehat{x_i}={\text {TPS}}(x,l,l'+\delta_l),   
\label{eq:tps}
\end{equation}
where ${\text {TPS}}$ and $\widehat{x_i}$ represent the spatial transformation and the deformed image noted as the intermediate face image. This face image has the appearance of $x_i$ and the landmarks of $x'_i$. The set $\delta_l$ accounts for small perturbations in the localizing the landmarks in the morph generation framework.
\subsection{Revisiting Landmarks and Appearance Disentanglement}

\begin{figure}
    \centering
    \includegraphics[width=240pt]{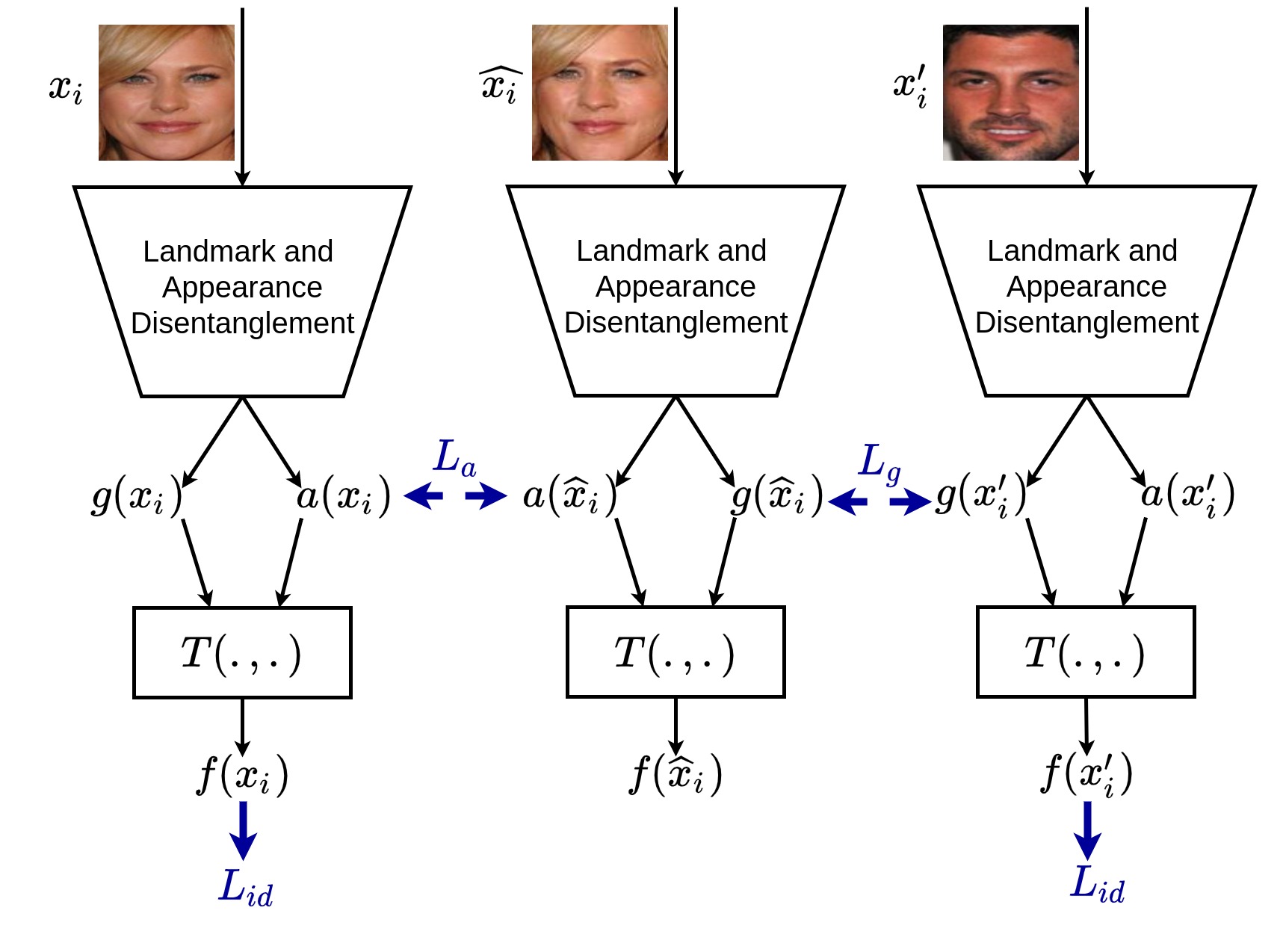}
    \caption{Face image $\widehat{x_i}$ is constructed by considering the appearance of $x_i$ and the landmarks of $x'_i$. $L_a$ enforces the appearance representations of $x_i$ and $\widehat{x_i}$ to be similar. Similarly, $L_g$ ensures that $g(\widehat{x_i})$ and $g({x'_i})$ are close to each other. A fully-connected layer of size $512$ fed with the concatenation of $g$ and $a$ provides the ID representation for the input image. 
    }
    \label{fig:firstexample_1}
\end{figure}

As presented in Figure~\ref{fig:firstexample_1}, in our proposed framework, two networks are defined as appearance network, $a$, and landmark network, $g$. These networks map the input face image to the appearance and landmark representations as: $a(.):\mathbb{R}^{w\times h\times 3}\rightarrow \mathbb{R}^{d_a}$ and $g(.):\mathbb{R}^{w\times h\times 3}\rightarrow \mathbb{R}^{d_g}$. 
It is worth mentioning that landmarks can be defined as the salient points in the face image. Although the landmark representation aims to represent the landmarks in the face image, it is trained through a classification setup to preserve the information required to distinguish between the input images regarding their geometrical differences. 
We define a third network, $f(.)$, that maps these two representations to a face ID representation as: $f(.):\mathbb{R}^{d_a}\times \mathbb{R}^{d_g}\rightarrow \mathbb{R}^{d_f}$, where $d_a$, $d_g$, and $d_f$ are the dimension of appearance, landmark, and face ID representations, respectively.  This representations enables us to train the framework as a classification setup.  

To provide enough information to distinguish between real and morph images, these three representations should satisfy three conditions: i) The appearance representation of the appearance and intermediate images should be similar: $a(x_i)\approx a(\widehat{x_i})$, ii) the landmark representation of the landmark and intermediate images should be similar:  $g(x'_i)\approx g(\widehat{x_i})$, and iii) for both the non-manipulated images, $x_i$ and $x'_i$, the face representations resulted from network $f$ should preserve sufficient classification information. We address these three conditions in our initial training setup. The appearance-preserving loss function aims to enforce the first condition: 
\begin{equation}
    L^1_a(x_i,\widehat{x_i})=-\frac{1}{N}\sum_i \Phi(a(x_i),a(\widehat{x_i})),
\end{equation}
where $\Phi(v_1,v_2)$ represents the cosine similarity between $v_1$ and $v_2$ as in~\cite{wen2016discriminative,liu2017sphereface}: $\Phi(v_1,v_2)=\frac{v_1^T v_2}{||v_1||_2 ||v_2||_2}$ and $N$ is the number of samples. Similarly, the landmark-preserving loss is defined as: 
\begin{equation}
\begin{aligned}
        L^1_g(x'_i,\widehat{x_i})=&-\frac{1}{N}\sum_i \Phi(g(x'_i),g(\widehat{x_i}))\\
        &+\textrm{max}(0,\Phi(g(x'_i),g({x_i}))-\alpha_g\phi_g),
\end{aligned}
\end{equation}
where $\phi_g=\frac{||l_i-l'_i||_2}{||l_i-\overline{l_i}||_2}$ is the normalized measure of the distance of landmark locations, and $\overline{l_i}$ is the mean of landmark locations along two axes. $\alpha_g$ is a scaling coefficient, scaling to form an angular loss which aims to maximize the cosine similarity of $g(x_i)$ and $g(\widehat{x_i})$ and dissimilarity 
of $g(x_i)$ and $g(x'_i)$.

In addition to the discussed training loss functions, we should assure that the appearance and landmark representations provide sufficient information for the identification of the real images, $x_i$ and $x'_i$: 
\begin{equation}
\small
    L^1_{id}(x_i)=\frac{-1}{N}\sum_i \log\frac{e^{s(\cos(m_1\theta_{y_i,i}+m_2)-m_3)}}{e^{s(\cos(m_1\theta_{y_i,i}+m_2)-m_3)}+\sum_{j\neq y_i}e^{s\cos(\theta_{j,i})}},
    \label{eq:arcface}
\end{equation}
where $f(x_i)=T(a(x_i),g(x_i))$ is the ID representation~\cite{deng2019arcface} for face image, $\cos(\theta_{j,i})=\frac{W_j^T f(x_i)}{||W_j||_2||f(x_i)||_2}$, and $W_j$ is the weight vector assigned to the $i^{th}$ class. In this angular loss function, $m_1$, $m_2$, and $m_3$ are the hyperparameters controlling the angular margin, and $s$ is the magnitude of angular representations. The training loss function is defined as:  
\begin{equation}
    L^1_t=\sum_i L^1_{id}(x_i)+L^1_{id}(x'_i)+\lambda^1_a L^1_a(x_i,\widehat{x_i})+\lambda^1_g L^1_g(x'_i,\widehat{x_i}),
\end{equation}
where $\lambda^1_a$ and $\lambda^1_g$ are hyper-parameters scaling the appearance and landmark preserving loss functions.

\begin{figure*}
    \centering
    \includegraphics[width=420pt]{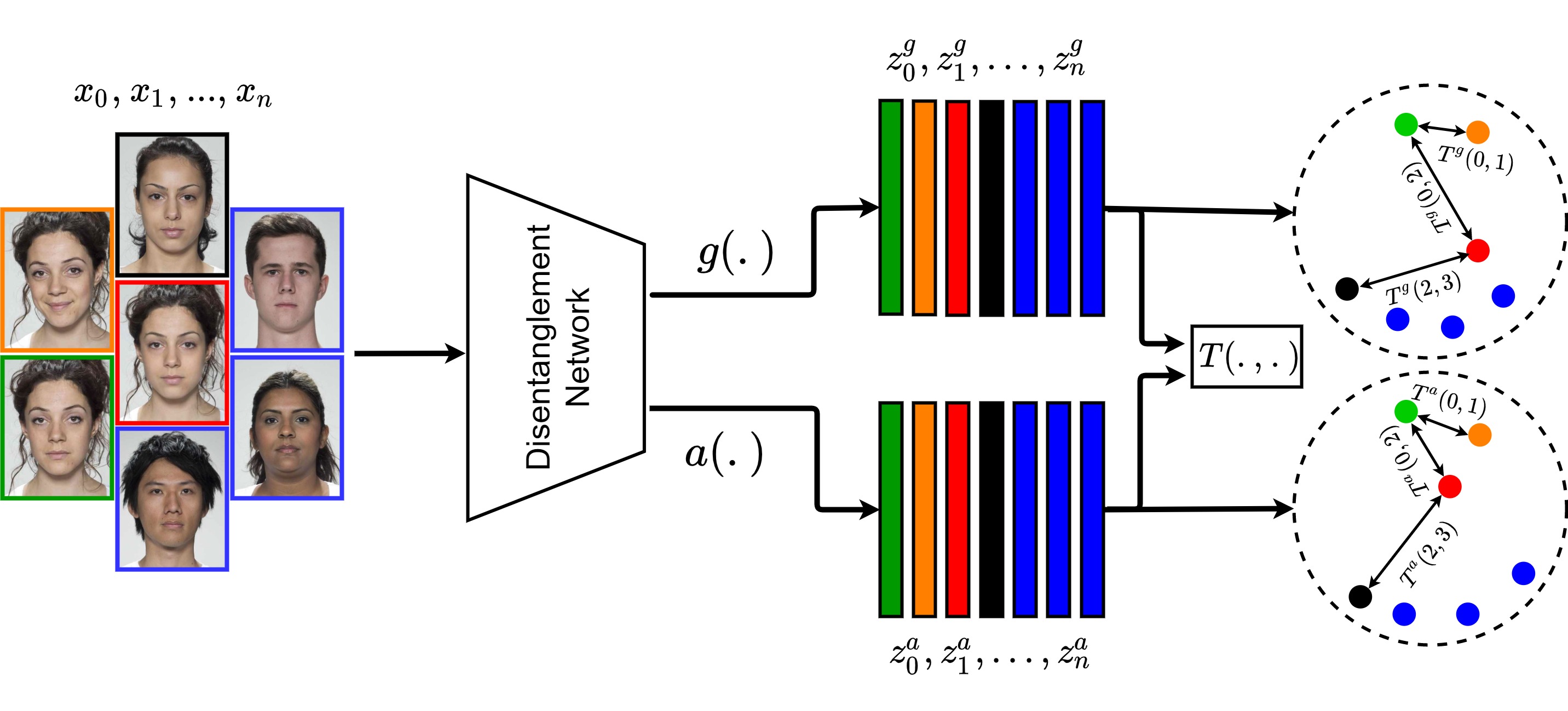}
    \caption{A pair of one trusted probe image, $x_i$, and an image in question, $x_j$, are fed into the disentanglement network. This network which is trained in combination with the auxiliary networks, $T^a(.,.)$ and $T^g(.,.)$, provides embedding representations that present high mutual information for genuine pairs and results in close representations for the samples in genuine pair and distant representations for samples in imposter pairs. Here, the morph image (red) is constructed displacing the landmarks of a real image (green) toward the landmarks of a visually similar image (black). The genuine pair consists of two real images from the same subject (orange and green), while the imposter pair in constructed using a real image and its corresponding morph image (green and red). 
    }
    \label{fig:contr}
\end{figure*}

\subsection{Contrastive Morph Detection}
Our proposed differential morph detection framework builds upon recent information-theoretic approaches to
deep representation learning~\cite{kemertas2020rankmi,tian2019contrastive}. We aim to maximize the mutual information between the real images from the same subject and minimize the mutual information between samples in an imposter pair during the training and make the decision during the test considering the distance between the representations of the pair of images in the embedding. To this aim, as presented in Figures~\ref{fig:contr}, the joint training of the disentanglement and auxiliary networks provides embedding representations distinguishable enough to detect morphed face images in a differential morph detection setup. Our framework benefits from transferring knowledge from that recognition task on a large face dataset to the disentanglement network, which provides a faster training of both disentanglement and auxiliary networks.

To maximize the mutual information between real samples from the same subject in the embedding space, we follow the notation proposed in~\cite{kemertas2020rankmi,belghazi2018mutual}. Let $x_i$ be an input face image and $z^a_i$ and $z^g_i$ be its corresponding appearance and landmark representations as: 
\begin{equation}
{z^a_i}=a(x_i), \\
{z^g_i}=g(x_i).
\end{equation}
We aim to train $a(x)$ and $g(x)$ such that real images from the same subject are mapped closely in the embedding space. To this aim, we maximize the mutual information between the real images from the same subject in each embedding space using the functions $T^a(.)$ and $T^g(.)$. To construct our training samples we define a genuine set as:

\begin{equation}
P=\{(x_i,x_j)|c_i=c_j, r_i=r_j=1\},
\label{positiveset}
\end{equation}
where $c_i$ and $c_j$ represents the classes for the subjects and $r_i=1$ represents the real images. On the other hand we define the imposter set as: 
\begin{equation}
N=\{(x_i,x_j)|c_i\neq c_j\;\textrm{or}\; r_i=0\;\textrm{or}\;r_j=0\},
\label{negativeset}
\end{equation}
where $r_i=0$ represents morphed images. It is worth mentioning that we define the above imposter set during the training. During the test phase, the imposer set consists of pairs in which both the samples belong to the same subject, while one of them is a real face image and the other is a morphed face image. In addition, for the genuine set, we can define the joint distribution of $x_i$ and $x_j$ as: 
\begin{equation}
p(x_i,x_j)=\sum_{k\in C} p(x_i,x_j,c=k, r_i=r_j=1).
\label{positiveprobability}
\end{equation}
Assuming the high entropy of $p(c)p(r)$ for the imposter set, we can approximate the joint distribution of the samples as the product of their marginals:  

\begin{equation}
\small
\begin{aligned}
&p(x_i)p(x_j)\approx\\ 
&\sum_{k\in c} \sum_{k'\neq k}\{p(x_i|c_i\!=\!k)p(x_j|c_j\!=\!k')p(c_i\!=\!k)p(c_j\!=\!k')\}\\
&+\sum_{k\in c} \sum_{r_i \in r}\{p(x_i|c_i\!=\!k)p(x_j|c_j\!=\!k)p(c_i\!=\!k)p(c_j\!=\!k)\\
&p(r_j\!=\!0)\}+\sum_{k\in c}  \{p(x_i|c_i\!=\!k)p(x_j|c_j\!=\!k')p(c_i\!=\!k)\\&\;\;\;p(c_j\!=\!k')p(r_i\!=\!0)p(r_j\!=\!1)\},
\end{aligned}
\label{negativeprobability}
\end{equation}
where $r=\{0,1\}$ represents morphed and real images.
Considering the genuine and imposter pairs defined in equations~\ref{positiveset} and~\ref{negativeset}, the appearance differential loss is defined to maximize the mutual appearance information between samples in a genuine pair as~\cite{belghazi2018mutual,kemertas2020rankmi}: 

\begin{equation}
\begin{aligned}
L^2_a=&\frac{1}{||P||}\sum_{(x_i,x_j)\in P}T^a(z^a_i,z^a_j)\\ -&\log \frac{1}{||N||} \sum_{(x_i,x_j)\in N} e^{T^a(z^a_i,z^a_j)}.
\end{aligned}
\label{trainingloss}
\end{equation}
A similar loss is defied over the genuine and imposter pairs to calculate $L^g_{t}$ as the differential landmark information loss. Then, the differential loss is defined as:      
\begin{equation}
\small
L^2_t=\lambda^2_a L^2_a+\lambda^2_g L^2_g+L^1_{id},
\label{diffloss}
\end{equation}
where $L^1_{id}$ provides the training for network $T$ and subsequently $f(x_i)$.

\section{Experiments}
We study the performance of the proposed framework on three morph datasets. In our experiments we follow frameworks described in~\cite{scherhag2020deep,venkatesh2020detecting}. 
Evaluation metrics for the differential morph detection are defined as: Attack Presentation Classification Error Rate (APCER) as the proportion of morph attack samples incorrectly classified as bona fide (non-morph), presentation and Bona Fide Presentation Classification Error Rate (BPCER) is the proportion of bona fide (nonmorph) samples incorrectly classified as morphed samples.

\subsection{Training Setup}
For all the datasets, DLib~\cite{king2009dlib} is considered to detect and align faces, as well as extracting $68$ landmarks. We train the model on the CASIA-WebFace~\cite{yi2014learning} dataset. In the training set, for each image, the image from a different ID that provides closest landmarks to its landmarks in terms of $L_2$ norm is selected. Neighbor face is transformed spatially using Equation~\ref{eq:tps}. This image is aligned again to compensate for the displacements caused by the spatial transformation. All images
are resized to $112 \times 112$ and pixel values are scaled to $[-1, 1]$.

We adopt ResNet-64~\cite{he2016deep} as the base network architecture. To reduce the size of the model, the convolutional networks for extracting the landmark representation, $g(x)$, and the appearance representation, $a(x)$, are combined. This network produces feature maps of spatial size $7 \times 7$ and the depth of $512$ channels. These feature maps are divided in depth into two sets, dedicated to the appearance and landmark representations, respectively. Each set of feature maps is reshaped to form a vector of size $12,544$ and passed to dedicated fully-connected layers. These layers of size $256$ generate the final representations, $a(x)$ and $g(x)$. The ID representation is constructed by concatenation of these two representations fed to a fully-connected layer of size $512$. The model is trained using Stochastic Gradient Descent
(SGD) with the mini-batch size of $128$ on two NVIDIA TITAN X GPUs. In Equation~\ref{eq:arcface}, following ArcFace~\cite{deng2019arcface} framework, $m_1$, $m_2$, and $m_3$ are set to 0.9, 0.4, and 0.15, respectively. In Equation~\ref{eq:tps}, $\delta_l$ is sampled from $N(0,3)$.

\begin{table*}[t]
\begin{center}
\begin{tabular}{l| c c c| c c c| c c c }
\hline
\multirow{2}{*}{{Dataset}}&\multicolumn{3}{c}{{MorGAN}}&\multicolumn{3}{c}{{VISAPP17}}&\multicolumn{3}{c}{{AMSL}}\\
&D-EER&5\%&10\%&D-EER&5\%&10\%&D-EER&5\%&10\%\\
\hline
LM-Dlib~\cite{damer2018detecting,king2009dlib}
                                &12.53& 20.71& 10.17&   17.88& 26.64&22.71&     14.45& 20.67&18.55\\
BSIF+SVM~\cite{kannala2012bsif} &10.17 &14.22& 8.64    &16.42 &28.77 &25.37&   12.75 &20.71&16.26\\
LBP+SVM~\cite{liao2007learning} &15.51& 28.40& 18.71    &18.75 &23.88 &20.65 &  14.97 &21.47&16.21\\
FaceNet~\cite{schroff2015facenet}&16.14&38.38&26.67     &9.51 &29.82 &6.91&     8.43 &25.74 &5.68\\
ArcFace~\cite{deng2019arcface}& 14.65&22.76& 16.23      &7.14&17.51&5.69 &      6.14&14.51&  5.23\\
FaceNet+SVM                   &12.53&18.84&12.21        &8.85 &26.46 &6.28     &8.42 &18.46 &5.28\\
ArcFace+SVM~\cite{scherhag2020deep}& 10.82 &15.47& 12.43    & 5.38 &7.45& 4.78         & 3.87 &6.12& 3.28\\
\hline
Ours& {\bf 8.75} &{\bf12.58}& {\bf8.51}         & {\bf4.69} &{\bf5.74}& {\bf2.59} & {\bf3.11} &{\bf5.35}& {\bf2.24}\\

\bottomrule
\end{tabular}
\end{center}
\caption[Table caption text]{D-EER\%, BPCER@APCER=5\%, and BPCER@APCER=10\% for the differential morph detection.}
\label{table:results_diff}
\end{table*}

The initial value for the learning rate is set to
$0.1$ and multiplied by $0.9$ in intervals of five epochs until its value is less than or equal to $10^{-6}$. The model is trained for $600$K iterations. We select $\alpha_g= 9.4$, $\lambda^1_a = 1.3$, and $\lambda^1_g = 0.75$. For training the network using Equation~\ref{trainingloss}, each fully-connected layer of size $256$ is fed to a fully-connected of size $64$, and then to a single unit. Here, considering $\lambda^2_a=\lambda^2_g=1$, the network is trained using the learning rate of $10^{-2}$ and is dropped similar to the rate mentioned above.

\subsection{Results}

\begin{figure}
    \centering
    \includegraphics[width=230pt]{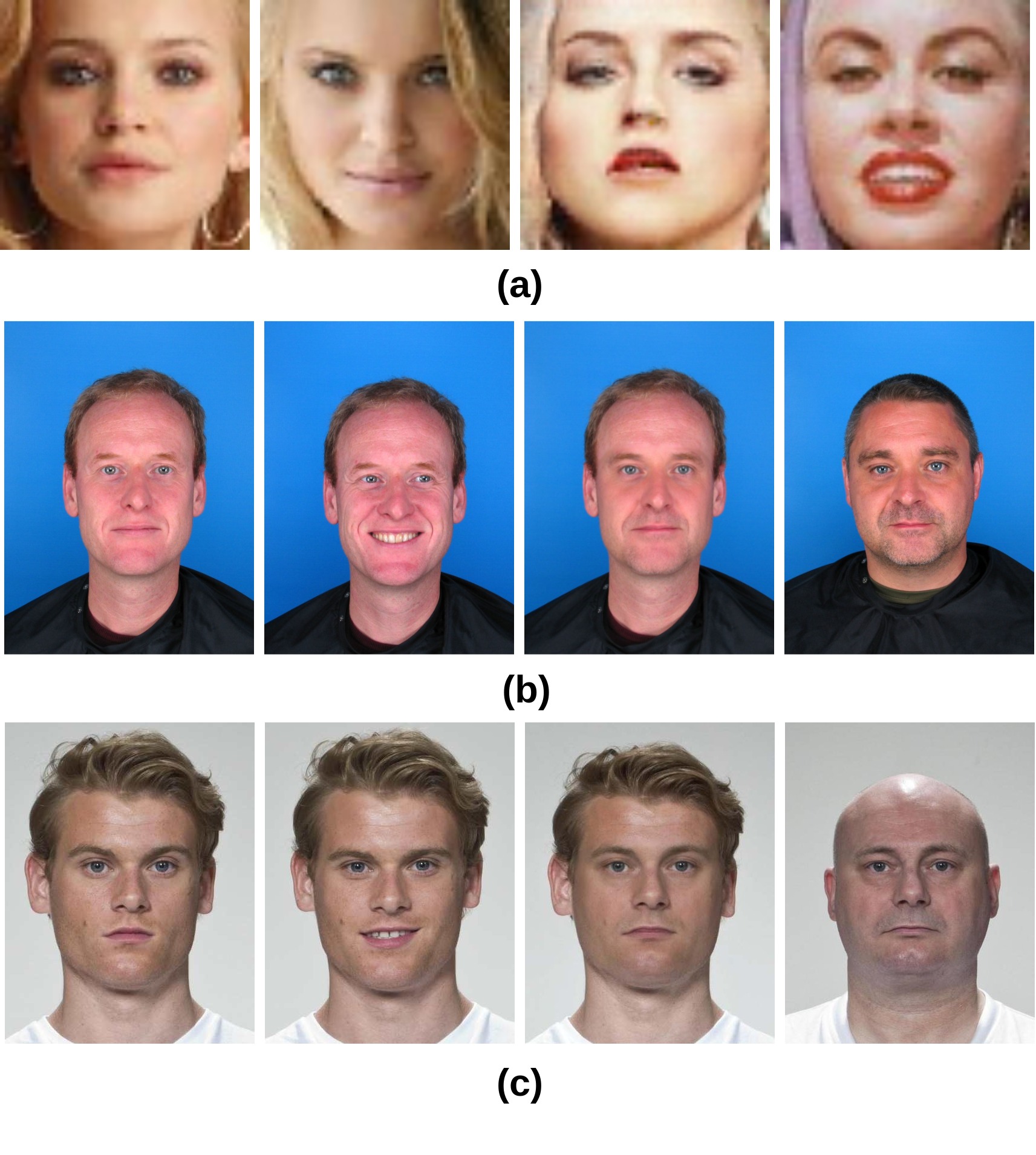}
    \caption{Samples from (a) MorGAN, (b) VISAPP17-Splicing-Selected, and (c) AMSL Face Morph Image Datasets. For each dataset, the first and second faces are the gallery and probe bona fide images and the third face is the morph image construed from the first and forth face images. The original sizes for face images in these datasets are $64\times 64$, $1500\times 1200$, and $531\times 413$, respectively. 
    }
    \label{fig:secondexample}
\end{figure}

\noindent{\bf MorGAN}  is constructed using the generative framework described in~\cite{damer2018morgan}. In this dataset, $500$ bonafide images are considered. For each bona fide image two morph images are generated using two most similar identities to the bona fide image, resulting in $1,000$ morph images. In total this dataset consists of $1,500$ references, $1,500$ probes, and $1,000$ MorGAN morphing attacks. The database is randomly split into disjoint and
equal train and test sets. All the images are of size $64\times 64$.

\noindent{\bf VISAPP17-Splicing-Selected}\footnote{For simplicity, we refer to this dataset as VISAPP17.} is a subset of VISAPP17-Splicing dataset~\cite{neubert2018extended} containing genuine neutral and smiling face images as well as morphed face images. This dataset is generated by warping and alpha-blending~\cite{wolberg1998image}.
To construct this dataset, facial landmarks are localized, the face image is tranquilized based on these landmarks, triangles are warped to some average position, and the resulting images are alpha-blended, where alpha is set to $0.5$ making alpha-blending equal to average. 
 Splicing morphs are
designed to avoid ghosting artefacts usually present in the hair
region, done by warping and blending of only face regions
and inserting the blended face into one of the original face images.
The background, hair and torso regions remain untouched. VISAPP17-Splicing-Selected dataset, which consists of $132$ bona fide and $184$ morph images of size $1500\times 1200$, is constructed by selecting morph images without any recognizable artifacts.

\noindent{\bf The AMSL Face Morph Image Dataset} is created using the Face Research Lab London Set~\cite{FaceResearchLabLondon} and includes genuine neutral and smiling face images and morphed face images. The morphed face images are generated from pairs of genuine face images~\cite{neubert2018extended}. For all the morph images the proportions of both faces in the morphed face are the same. While generating morphed faces male, female, white, and Asian people are only morphed with their corresponding category. All images are down-scaled to $531\times 413$ pixels and JPEG compression is applied to them to compress the images to 15kb~\cite{wolf2016portrait}. This dataset includes 102 neutral or 102 smiling genuine face images and 2,175 morph images.

\noindent{\bf Differential Morph Detection:} For the MorGAN dataset, we follow the train and test split presented in~\cite{damer2018morgan}. For the other two datasets, we consider a disjoint train and test split in which $50\%$ of the subjects are used for training. The distance between face images $x_i$ and $x_j$ is defined as: 
\begin{equation}
\begin{aligned}
        D=&\Phi(f(x_i),f(x_j))+\beta_a\Phi(a(x_i),a(x_j))\\+&\beta_g\Phi(g(x_i),g(x_j)),
\end{aligned}
\label{decison}
\end{equation}
where $\beta_a$ and $\beta_g$ are the scaling parameters used for decision making. We employ classical texture descriptors, BSIF~\cite{kannala2012bsif} and
LBP~\cite{liao2007learning}, with an SVM
classifier. The LBP feature descriptors are extracted according to the original LBP image patches of $3\times3$. The resulting feature vector is then a normalized histogram of size 256, which encompasses all
potential values of the LBP binary code. BSIF feature vectors are conducted on a filter size of $3\times3$ and 8 bits. The filters utilized for BSIF are pre-learned Independent
Component Analysis (ICA) filters~\cite{hyvarinen2009natural} that are utilized by the original BSIF paper to construct normalized histogram for each image. The feature vectors are then inputted 
to an SVM with an RBF kernel for classification. For all classical baseline models the difference between the feature representation of the image in question and the feature representation of the trusted image is fed to an SVM classifier.  
\begin{table}[t]
\small
\begin{center}
\addtolength{\tabcolsep}{-0pt}
\begin{tabular}{lccccc}
\hline
Train&Test&Algorithm&D-EER&5\%&10\%\\ \hline
\multirow{8}{*}{\rotatebox[origin=c]{90}{MorGAN}}&\multirow{4}{*}{\rotatebox[origin=c]{90}{VISAPP17}}&LM-Dlib~\cite{damer2018detecting,king2009dlib}&                      23.74 &51.42&38.67\\
                                &&BSIF+SVM~\cite{kannala2012bsif}&     19.21 &51.25&39.41\\
                                &&ArcFace+SVM~\cite{scherhag2020deep}&  11.67 &22.36 & 14.86\\
                                &&Ours&                                 {\bf8.55} &{\bf12.68}&{\bf8.57}\\\cline{2-6}
                                
&\multirow{4}{*}{\rotatebox[origin=c]{90}{AMSL}}&LM-Dlib~\cite{damer2018detecting,king2009dlib}
                                                                &      20.67 &44.28&32.15\\
                                &&BSIF+SVM~\cite{kannala2012bsif}&     17.27 &38.54&24.71\\
                                &&ArcFace+SVM~\cite{scherhag2020deep}& 10.48 &22.49 &14.90\\
                                &&Ours&                                 {\bf7.95} &{\bf11.26} &{\bf8.81}\\\hline
                                
\multirow{8}{*}{\rotatebox[origin=c]{90}{VISAPP17}}&\multirow{4}{*}{\rotatebox[origin=c]{90}{MorGAN}}&LM-Dlib~\cite{damer2018detecting,king2009dlib}                                    &16.82& 38.54& 24.8\\
                                &&BSIF+SVM~\cite{kannala2012bsif}&{\bf{13.52}} &15.3& 14.79\\
                                &&ArcFace+SVM~\cite{scherhag2020deep}& 15.75 &32.58& 22.36\\
                                &&                                      Ours& 13.85 &{\bf 12.32}& {\bf8.74}\\\cline{2-6}
                                
&\multirow{4}{*}{\rotatebox[origin=c]{90}{AMSL}}&LM-Dlib~\cite{damer2018detecting,king2009dlib}
                                                                     &18.83& 38.86&24.78\\
                                &&BSIF+SVM~\cite{kannala2012bsif}    &16.92 &38.84&24.64\\
                                &&ArcFace+SVM~\cite{scherhag2020deep}& 8.27 &9.63& 5.28\\
                                &&Ours                               & {\bf5.38} &{\bf3.47}& {\bf2.38}\\\hline
                                
\multirow{8}{*}{\rotatebox[origin=c]{90}{AMSL}}&\multirow{4}{*}{\rotatebox[origin=c]{90}{MorGAN}}&LM-Dlib~\cite{damer2018detecting,king2009dlib}                                        &16.24& 30.94& 19.28\\
                                &&BSIF+SVM~\cite{kannala2012bsif}&{\bf13.84} &{\bf25.35}& {\bf14.82}\\
                                &&ArcFace+SVM~\cite{scherhag2020deep}& 16.34 &38.62& 24.51\\
                                &&Ours& 14.21 &28.58& 18.51\\\cline{2-6}
                                
&\multirow{4}{*}{\rotatebox[origin=c]{90}{VISAPP17}}&LM-Dlib~\cite{damer2018detecting,king2009dlib}                                                                                    & 20.55& 62.21&38.42\\
                                &&BSIF+SVM~\cite{kannala2012bsif}&20.36 &51.28 &32.95\\
                                &&ArcFace+SVM~\cite{scherhag2020deep}& 10.65 &14.36& 9.81\\
                                &&Ours                               & {\bf5.21} &{\bf8.26}& {\bf4.17}\\\hline
\end{tabular}
\end{center}
\caption[Table caption text]{Cross-dataset performance for differential morph detection: D-EER\%, BPCER@APCER=5\%, and BPCER@APCER=10\%.}       
\label{table:results_cross}
\end{table}

\begin{table*}[t]
\begin{center}
\begin{tabular}{l| c c c| c c c| c c c }
\hline
\multirow{2}{*}{{Dataset}}&\multicolumn{3}{c}{{MorGAN}}&\multicolumn{3}{c}{{VISAPP17}}&\multicolumn{3}{c}{{AMSL}}\\
&D-EER&5\%&10\%&D-EER&5\%&10\%&D-EER&5\%&10\%\\
\hline
LM-Dlib~\cite{damer2018detecting,king2009dlib}
            &8.14& 10.67 &7.83                             &15.67 &22.87 &20.32 &11.67 & 16.98 &14.63\\
BSIF+SVM~\cite{kannala2012bsif}  &6.07 &9.15 &4.63          &13.87 &23.53 &20.12 &10.53 &16.53  &13.86\\
LBP+SVM~\cite{liao2007learning}  &7.47 &9.23& 4.71          &15.21 &20.64 &18.74 &12.21 &17.11  &12.81  \\
FaceNet~\cite{schroff2015facenet}&8.11&14.52 &7.59         &7.32  &24.54 &5.21  &7.46  &22.12  &5.17\\
ArcFace~\cite{deng2019arcface}   & 7.58&   9.64& 4.08        &6.45  &14.78 &5.02  &5.36  &10.46  &4.87  \\
FaceNet+SVM                      &7.23&12.46& 5.22          &6.37  &26.46 & 6.28  &8.42  &18.46  &5.28\\
ArcFace+SVM~\cite{scherhag2020deep} & 5.35 &6.71 &3.50    &4.52  &5.98  &4.05  &3.27  &5.56   & 2.69\\
\hline
Ours& 4.71 &5.32& 3.85         & 3.74 &4.91& 2.17       & 2.82 &4.97& 2.82\\
Ours$^*$& {\bf4.06} &{\bf5.04}& {\bf3.42}         & {\bf3.45} &{\bf4.25}& {\bf1.85}       & {\bf2.36} &{\bf4.16}& {\bf1.47}\\

\bottomrule
\end{tabular}
\end{center}
\caption[Table caption text]{The differential morph detection performance on three datasets, when the trusted image is known to the detection framework: D-EER\%, BPCER@APCER=5\%, and BPCER@APCER=10\%.}
\label{table:results_trusted}
\end{table*}

In addition, we employ LM-Dlib~\cite{damer2018detecting,king2009dlib} as a model for the landmark displacement measure. In this framework, the distance between landmarks extracted by Dlib~\cite{king2009dlib} are fed to an SVM. For deep models, the distance between the representations in the embedding domain is considered as the decision criteria. For all the model, the default parameters presented in the original papers are considered. It is worth mentioning that in this experiments we do not consider the prior knowledge on which of the images in the pair fed to the recognition framework is the trusted image. On the other hand, in Table~\ref{table:results_trusted}, we assume that the differential morph detection framework is provided with the information regarding the trusted image.  

\begin{figure}
    \centering
    \includegraphics[width=190pt]{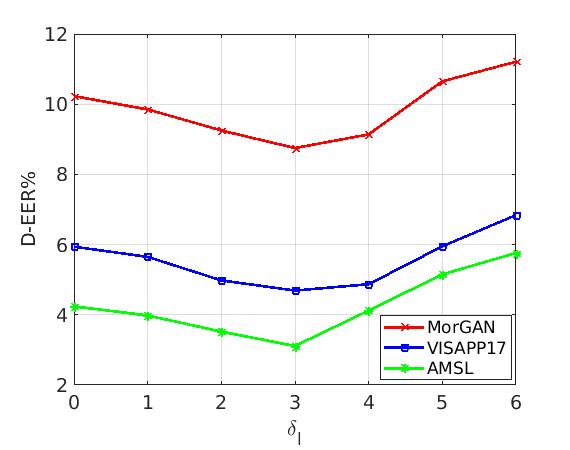}
    \caption{D-EER\% for different variances of $\delta_l$ values in Equation~\ref{eq:tps}.
    }
    \label{fig:delta}
\end{figure}

For each the datasets, $10\%$ of the training set is considered as the validation set. Then, the parameters to train the framework are selected based on the experiments described in Table~\ref{table:results_beta} and Figure~\ref{fig:delta}. Table~\ref{table:results_diff} presents the performance of the proposed framework in comparison with four deep learning and three classical differential morph detection frameworks. In addition to outperforming the baseline models on all the datasets, the proposed framework outperforms the baseline models by a wide margin on the MorGAN dataset, which can be contributed to the disentanglement of landmark and appearance representations.

In Table~\ref{table:results_cross}, we study the performance of the networks trained on the training portion of one morph dataset and tested on the other datasets. As presented in this table, while outperforming the other models, the proposed framework provides high cross-dataset performance between VISAPP17 and AMSL. In addition, the proposed framework provides D-EER of 8.55\% and 7.95\% for cross-dataset performance on the network trained on MorGAN and tested on VISAPP17 and AMSL datasets, respectively. On the other hand, BSIF+SVM outperforms the other algorithms when testing the network trained on other two datasets and tested on MorGAN, which illustrates the same trend as the results provided in~\cite{damer2018morgan}.

Table~\ref{table:results_trusted} studies the effect of the trusted images being known to the detection framework. For the baseline models, rather than comparing the representations of the trusted images and the image in question, the representation of the image in question is subtracted from the representation of the trusted image before feeding the difference to the SVM. For the proposed framework, we consider an additional algorithm, denoted as "Ours$^*$", in which two dedicated instances of the framework are constructed for trusted images and images in question. In this algorithm, which outperforms the algorithm for which only one instance of the network is considered, we only train the network dedicated to the images in question. Table~\ref{table:results_beta} provides the performance for the proposed framework on the validation sets when the scaling parameters in making the decision vary in Equation~\ref{decison}. As presented in this table, morph images constructed using landmark displacement are better detected for higher weights given to $g(x)$, while the MorGAN samples are best detected when $g(x)$ and $a(x)$ are given similar weights. In addition, Figure~\ref{fig:delta} provides the performance for three datasets when variance of the normal distribution to generate $\delta_l$ samples in Equation~\ref{eq:tps} varies from 0 to 6. 

\begin{table}[t]
\begin{center}
\begin{tabular}{l| c c c}
\hline
&{{MorGAN}}&{{VISAPP17}}&{{AMSL}}\\
\hline
(4,1)&10.91       &6.97       &   4.72\\
(3,1)&9.64       &6.57       & 4.12\\
(2,2)&{\bf{8.75}}&5.84       & 3.83\\
(1,3)&10.32       &{\bf{4.69}}& {\bf{3.11}}              \\
(1,4)&10.89       &5.12&3.54\\
\bottomrule
\end{tabular}
\end{center}
\caption[Table caption text]{The D-EER\% for differential morph detection performance considering different scaling values ($\beta_a$ and $\beta_g$) in Equation~\ref{decison}. }
\label{table:results_beta}
\end{table}

\section{Conclusions}
In this paper, we presented a novel differential morph detection framework which benefits from disentangling landmark representation and appearance representation in an embedding space. These two representations which are disentangled but complementary, are constructed using a disentanglement network trained using triplets of face images. Each triplet consists of two real images and an intermediate image which inherits the landmarks from one image and the appearance from the other image. We demonstrated that appearance and landmark disentanglement can be boosted using contrastive representations for each disentangled representation. This property provides the possibility of accurate differential morph detection, using distances in landmark, appearance, and ID domains. The performance of the proposed framework is studied using three morph datasets constructed with different methodologies.

{\small
\bibliographystyle{ieee_fullname}
\bibliography{egbib}
}

\end{document}